\documentclass{bmvc2k}

\usepackage{multirow}
\usepackage{booktabs}
\usepackage{colortbl}
\definecolor{mygray}{gray}{0.95}

\title{Diagnosing Errors in Video Relation Detectors}

\addauthor{Shuo Chen \\
Pascal Mettes \\
Cees G.M. Snoek
}{{s.chen3, p.s.m.mettes, cgmsnoek}@uva.nl}{1}

\addinstitution{
VIS Lab, \\
University of Amsterdam, \\
The Netherlands}

\runninghead{Chen, Mettes, Snoek}{Diagnosing Errors in Video Relation Detectors}

\def\etal{\emph{et al}\bmvaOneDot}

\usepackage{caption, subcaption,colortbl}
\usepackage{multirow}
\usepackage{graphicx}
\usepackage{booktabs,amsfonts,dcolumn,enumitem}

\newcommand{\ie}{\textit{i}.\textit{e}., }
\definecolor{Gray}{gray}{0.9}
\definecolor{Grayer}{gray}{0.95}

\begin{document}

\maketitle

\begin{abstract}
Video relation detection forms a new and challenging problem in computer vision, where subjects and objects need to be localized spatio-temporally and a predicate label needs to be assigned if and only if there is an interaction between the two. Despite recent progress in video relation detection, overall performance is still marginal and it remains unclear what the key factors are towards solving the problem. Following examples set in the object detection and action localization literature, we perform a deep dive into the error diagnosis of current video relation detection approaches. We introduce a diagnostic tool for analyzing the sources of detection errors. Our tool evaluates and compares current approaches beyond the single scalar metric of mean Average Precision by defining different error types specific to video relation detection, used for false positive analyses. Moreover, we examine different factors of influence on the performance in a false negative analysis, including relation length, number of subject/object/predicate instances, and subject/object size. Finally, we present the effect on video relation performance when considering an oracle fix for each error type. On two video relation benchmarks, we show where current approaches excel and fall short, allowing us to pinpoint the most important future directions in the field. The tool is available at \url{https://github.com/shanshuo/DiagnoseVRD}.
\end{abstract}

\section{Introduction}

This paper performs an in-depth investigation into the video relation detection task. Video relation detection, introduced by Shang~\etal~\cite{shang2017video}, requires spatio-temporal localization of object and subject pairs in videos, along with a predicate label that describes their interaction. 
To tackle this challenging problem, Shang~\etal~\cite{shang2017video} first proposed a three-stage approach: split a video into snippets, predict the predicate, and associate the snippets over time. Such a three-stage tactic has since become popular for video relation detection~\cite{di2019multiple,tsai2019video,qian2019video,su2020video,xie2020video}. Among them, Tsai~\etal~\cite{tsai2019video}, Qian~\etal~\cite{qian2019video} and Xie~\etal~\cite{xie2020video} focus on improving predicate prediction. Tsai~\etal and Qian~\etal construct graphs to pass messages between object nodes, while Xie~\etal utilizes multi-modal features. Alternatively, both Di~\etal~\cite{di2019multiple} and Su~\etal~\cite{su2020video} shift their attention to a better association process.

Not all works follow a canonical three-stage approach. Cao~\etal~\cite{cao2021relation}, for example,  propose a 3D proposal network to learn relational features in an end-to-end manner.
Sun~\etal~\cite{sun2019video} and Liu~\etal~\cite{liu2020beyond} rely on a sliding window to generate proposals and recognize predicates within proposals.
Chen~\etal~\cite{chen2021social} learn interaction primitives to generate interaction proposals~\cite{chen2020interactivity} and recognize predicates.
While video relation results keep progressing, there is still a lot of room for improvement. For example, Xie~\etal~\cite{xie2020video}, the winner of the Video Relation Detection task from the Video Relation Understanding Challenge 2020, combines a wide variety of multi-modal features for each subject-object tubelet pair to predict the relations with an improved detection performance. Nonetheless, their final mAP (mean Average Precision) is only 9.66\% on the VidOR validation set~\cite{shang2019annotating}.
In short, the task is far from solved.
Moreover, it is unclear which factors are most critical for better results. We seek to fill this void.

We take inspiration from error diagnosis in the spatial domain for object detection~\cite{hoiem2012diagnosing,bolya_2020eccvtide} and in the temporal domain for action detection~\cite{alwassel2018diagnosing,otani_bmvc2020uncover}. 
These works have previously performed a deep dive into the main sources of errors for their respective tasks, including false positive analysis, false negative analysis, and mAP sensitivity tests for object attributes or action characteristics. The analyses have helped to explain limitations in the field and to provide guidance for the next steps~\cite{hoiem2012diagnosing,hosang2014good,benenson2014ten,hosang2015makes, bmvc2015diagnosing,zhang2016far,agrawal2016analyzing,alwassel2018diagnosing,bolya_2020eccvtide,fan2019tracklinic}. In a similar spirit, we shine a light on the spatio-temporal domain for video relation detection, where the spatial challenges of object detection and the temporal challenges of action detection need to be simultaneously addressed.

We provide an error diagnosis for video relation detection, which starts with an outline of current benchmarks, evaluation protocols, the algorithms under consideration, and a categorisation of different possible error types. Under this setup, we make the following analytical contributions:
\begin{itemize}[noitemsep]
\item false positive analysis outlining which types of errors are most common, along with potential cures for each error type, evaluated on two state-of-the-art approaches;
\item false negative analysis along with a categorization of the kind of relation characteristics that are most difficult to detect;
\item analysis of the different video relation characteristics and their influence on the performance, including relation length, number of subject/object/predicate instances, and spatio-temporal subject and object size;
\item oracle analysis to identify which aspects lead to the biggest improvements.
\end{itemize}

\section{Error diagnosis setup}
As a starting point of the error diagnosis, we first outline the core characteristics and biases of the current video relation detection datasets, the definitions of different error types, and the methods from the literature under investigation.

\subsection{Dataset characterization}
We perform our analysis on the two existing datasets in video relation detection, namely ImageNet-VidVRD~\cite{shang2017video} and VidOR~\cite{shang2019annotating}.

\textbf{ImageNet-VidVRD} \cite{shang2017video} consists of 1,000 videos, created from the ILSVRC2016-VID dataset \cite{russakovsky2015imagenet}. There are 35 object categories and 132 predicate categories. The videos are densely annotated with relation triplets in the form of \emph{⟨subject-predicate-object⟩} as well as the corresponding subjects and objects trajectories. Following \cite{shang2017video, tsai2019video}, we use 800 videos for training and the remaining 200 for testing. We analyze the method performance on the 200 test videos.

\textbf{VidOR} \cite{shang2019annotating} contains 10,000 user-generated videos selected from YFCC-100M \cite{thomee2016yfcc100m}, for a total of about 84 hours. There are 80 object categories and 50 predicate categories. Besides providing annotated relation triplets, the dataset also provides the bounding boxes of objects. The dataset is split into a training set with 7,000 videos, a validation set with 835 videos, and a testing set with 2,165 videos. Since the ground truth of the test set is not available, we use the training set for training and the validation set for testing, following \cite{liu2020beyond,qian2019video, xie2020video, su2020video}. We report the analysis of method performance on the VidOR validation set.

\begin{figure}
\captionsetup{aboveskip=5pt}
\captionsetup[subfigure]{aboveskip=0pt}
\centering
\begin{subfigure}{0.39\textwidth}
\centering
\includegraphics[width=\linewidth]{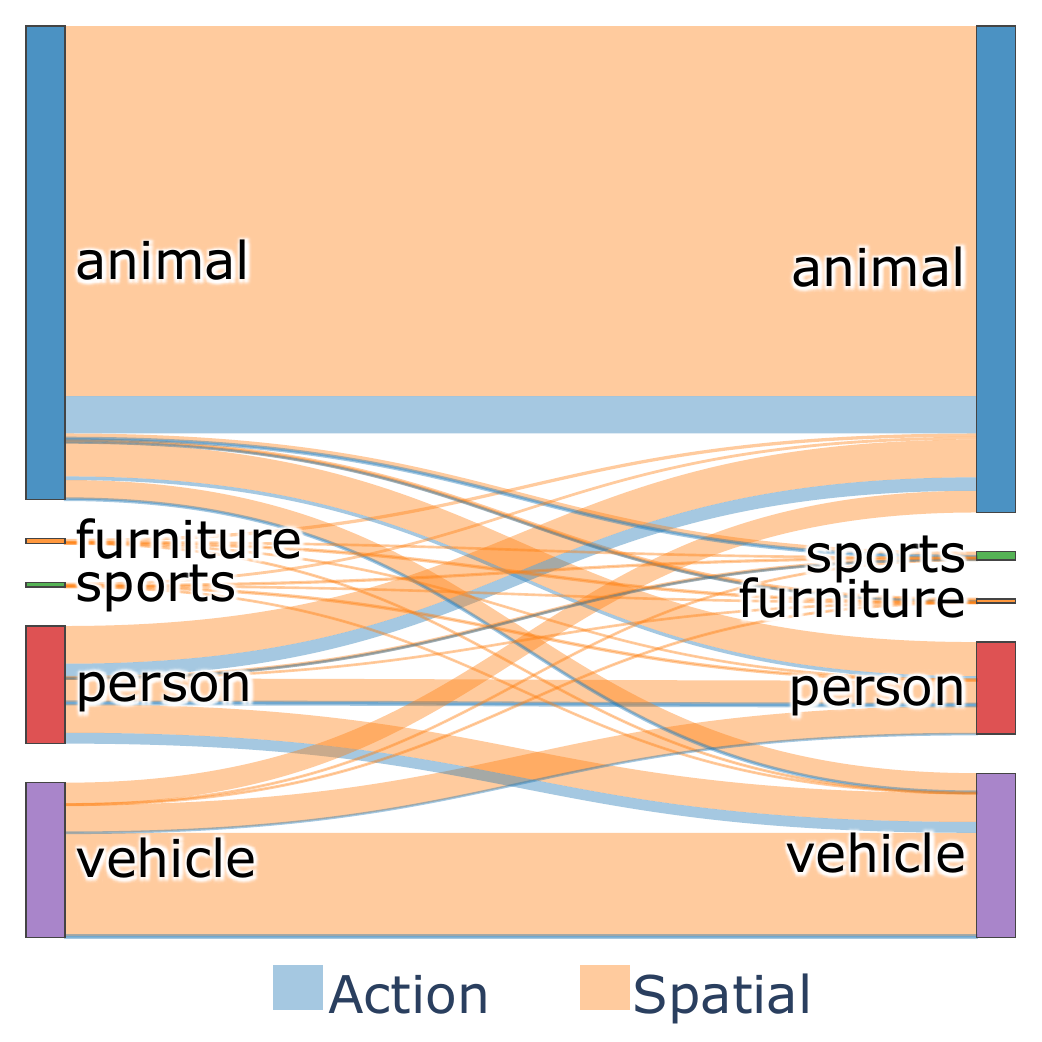}
\caption{ImageNet-VidVRD.}
\end{subfigure}
\hspace{0.25cm}
\begin{subfigure}{0.39\textwidth}
\centering
\includegraphics[width=\linewidth]{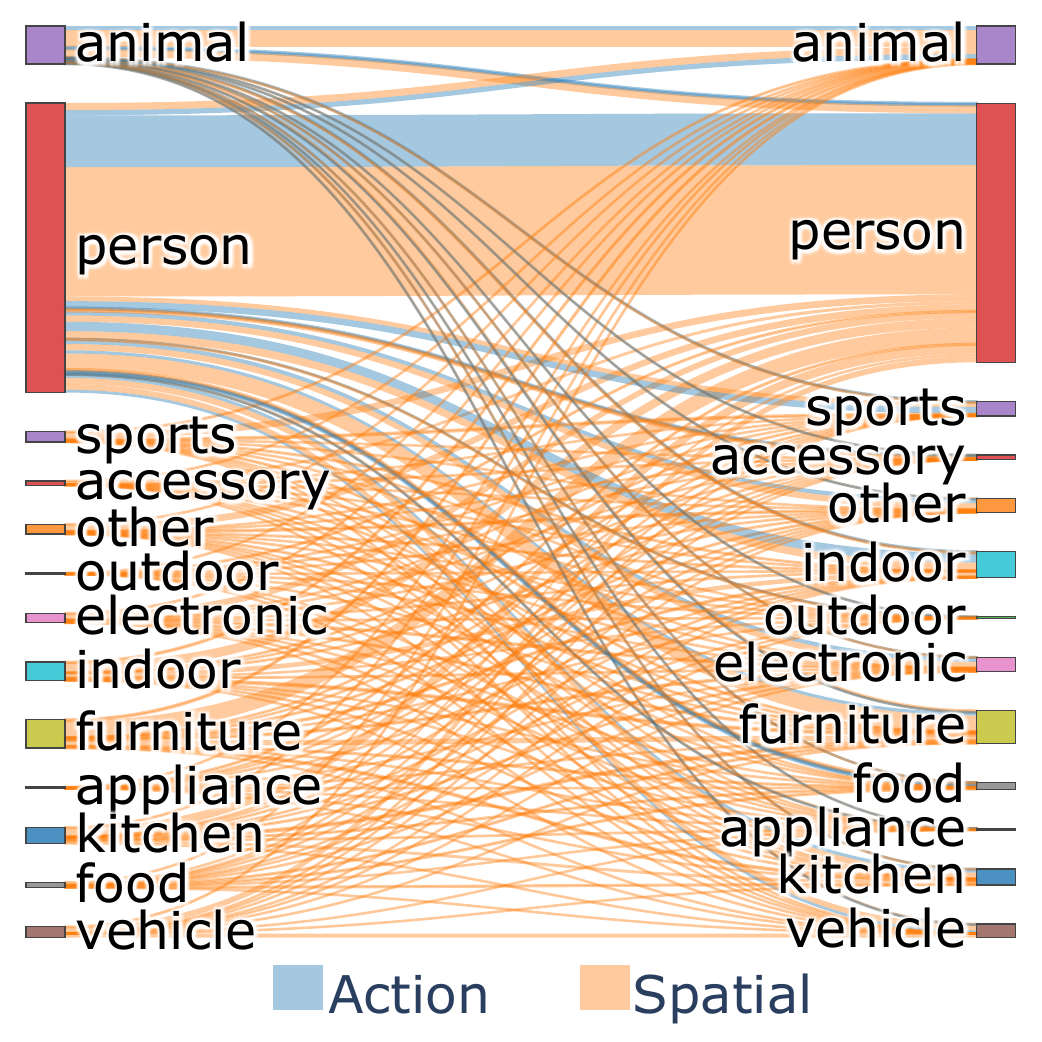}
\caption{VidOR.}
\end{subfigure}
  \caption{
Subject, object, and predictate diagrams on ImageNet-VidVRD and VidOR. On both datasets, knowledge about animals, person, vehicles, and spatial relations will go a long way for video relation detection due to a large bias towards these overarching category types. 
  }
\label{fig:relations}
\end{figure}

\textbf{Prevalent relations.}
To gain insight into the large number of possible combinations of subjects, objects, and interactions in ImageNet-VidVRD and VidOR, we first categorize all into super categories and investigate patterns among the super categories. 
For VidOR, the object categories are based on MS-COCO~\cite{lin2014microsoft} and we, therefore, use its 12 object super categories, along with an \textit{other} category for exceptions. 
For the predicates, we employ the hierarchy in VidOR that makes a split into \textit{action-based} and \textit{spatial} predicates.
In the supplementary materials, we show the
prevalent objects and predicates of ImageNet-VidVRD and VidOR. The animals and persons are the dominant subjects and objects, while spatial predicates form the dominant interactions between them. This is not surprising, as spatial relations are common and omnipresent.

\textbf{Predicate biases.}
For a given dataset, the number of relations consists of all combinations of subjects, objects, and predicates. Most combinations are however not likely to occur, resulting in a bias towards common and generic \emph{⟨subject-predicate-object⟩} triplets.
We find that subject and object labels are highly predictive of predicate labels. Figure~\ref{fig:relations} shows which subjects and objects are likely to be in interaction and indicates which type of predicate commonly occurs between super categories of subjects and objects. To quantify the bias towards predicate categories for subject-object pairs, we predict the predicate using a na\"ive Bayes classifier built upon training set statistics between subjects and objects. On ImageNet-VidVRD, the predicate accuracy on the validation set is 14.02\% compared to 0.8\% for random guessing. On VidOR, the accuracy is 36.11\% compared to 2.0\% for random guessing. Evidently, there is not only a strong bias towards common predicates but also from subjects and objects to predicates. Empirically, we will investigate whether current video relation detection approaches also mirror this bias.

\begin{table}[t!]
\centering
\captionsetup{aboveskip=5pt}
\resizebox{0.875\linewidth}{!}{%
\begin{tabular}{lll}
\toprule
Error type & Definition\\
\midrule
\textbf{Classification error} & \begin{tabular}{@{}l@{}}Overlap between discovered and ground truth relation\\is above 0.5, the relation triplet labels are not identical. \end{tabular}\\
\rowcolor{Grayer}
\textbf{Localization error} & \begin{tabular}{@{}l@{}}Overlap between discovered and ground truth relation is\\between 0.1 and 0.5, the relation triplets labels are identical.\end{tabular}\\
\textbf{Confusion error} & \begin{tabular}{@{}l@{}}Overlap between discovered and ground truth relation is\\between 0.1 and 0.5, the relation triplets are not identical.\end{tabular}\\
\rowcolor{Grayer}
\textbf{Double detection} & \begin{tabular}{@{}l@{}}Overlap between discovered and ground truth relation is\\above 0.5, the relation triplet are identical, but the\\ground truth instance has already been detected.\end{tabular}\\
\textbf{Background error} & \begin{tabular}{@{}l@{}}Overlap between discovered and \emph{any}\\ground truth relation is lower than 0.1.\end{tabular}\\
\rowcolor{Grayer}
\textbf{Missed ground truth} & \begin{tabular}{@{}l@{}}An undetected ground truth instance\\not covered by other errors.\end{tabular}\\
\bottomrule
\end{tabular}
}
\caption{
Categorization of six different types covering all errors that a video relation detector can make. The error types are used for our in-depth false positive analysis.
}
\label{tab:errortypes}
\end{table}

\subsection{Evaluation protocol and error types}
In the literature, the mean Average Precision (mAP) is widely used for video relation detection evaluation~\cite{shang2017video, shang2019relation,tsai2019multimodal, qian2019video,sun2019video,su2020video,liu2020beyond,xie2020video}.
Different from conventional Average Precision evaluation for detection~\cite{everingham2010pascal}, the averaging per category is performed over videos, not categories.
Let $G$ be the set of ground truth instances for a video such that an instance $g^{(k)} {=} ({\langle s,p,o \rangle}^g, (T_s^g, T_o^g))$ consists of a relation triplet label ${\langle s,p,o \rangle}^g$ with subject and object bounding-box trajectories $(T_s^g, T_o^g)$. Let $P$ be the set of predictions such that a prediction $p^{(i)} {=} (p_s^{(i)}, {\langle s,p,o \rangle}^p, (T^g_s,T^g_o)$ consists of a relation triplet score $p_s^{(i)}$, a triplet label ${\langle s,p,o \rangle}^p$, and predicted subject and object trajectories. To match a predicted relation instance ($\langle s,p,o \rangle^p, (T^p_s,T^p_o)$) to a ground truth ($\langle s,p,o \rangle^g, (T^g_s,T^g_o$)), we require:
\begin{enumerate}[label=\roman*,noitemsep]
    \item their relation triplets to be exactly the same, i.e. $\langle s,p,o \rangle^p {=} \langle s,p,o \rangle^g$;
    \item their bounding-box trajectories overlap s.t. vIoU$(T^p_s,T^g_s) \geq 0.5$ and vIoU$(T^p_o,T^g_o) \geq 0.5$, where vIoU refers to the voluminal Intersection over Union;
    \item the minimum overlap of the subject trajectory pair and the object trajectory pair $\text{ov}_{pg} = \text{min}(\text{vIoU}(T^p_s,T^g_s), \text{vIoU}(T^p_o,T^g_o))$ is the maximum among those paired with the other unmatched ground truths $G$, \ie $\text{ov}_{pg} \geq \text{ov}_{pg'}(g' \in G)$.
\end{enumerate}

\begin{figure}
\captionsetup{aboveskip=5pt}
    \centering
    \includegraphics[width=\linewidth]{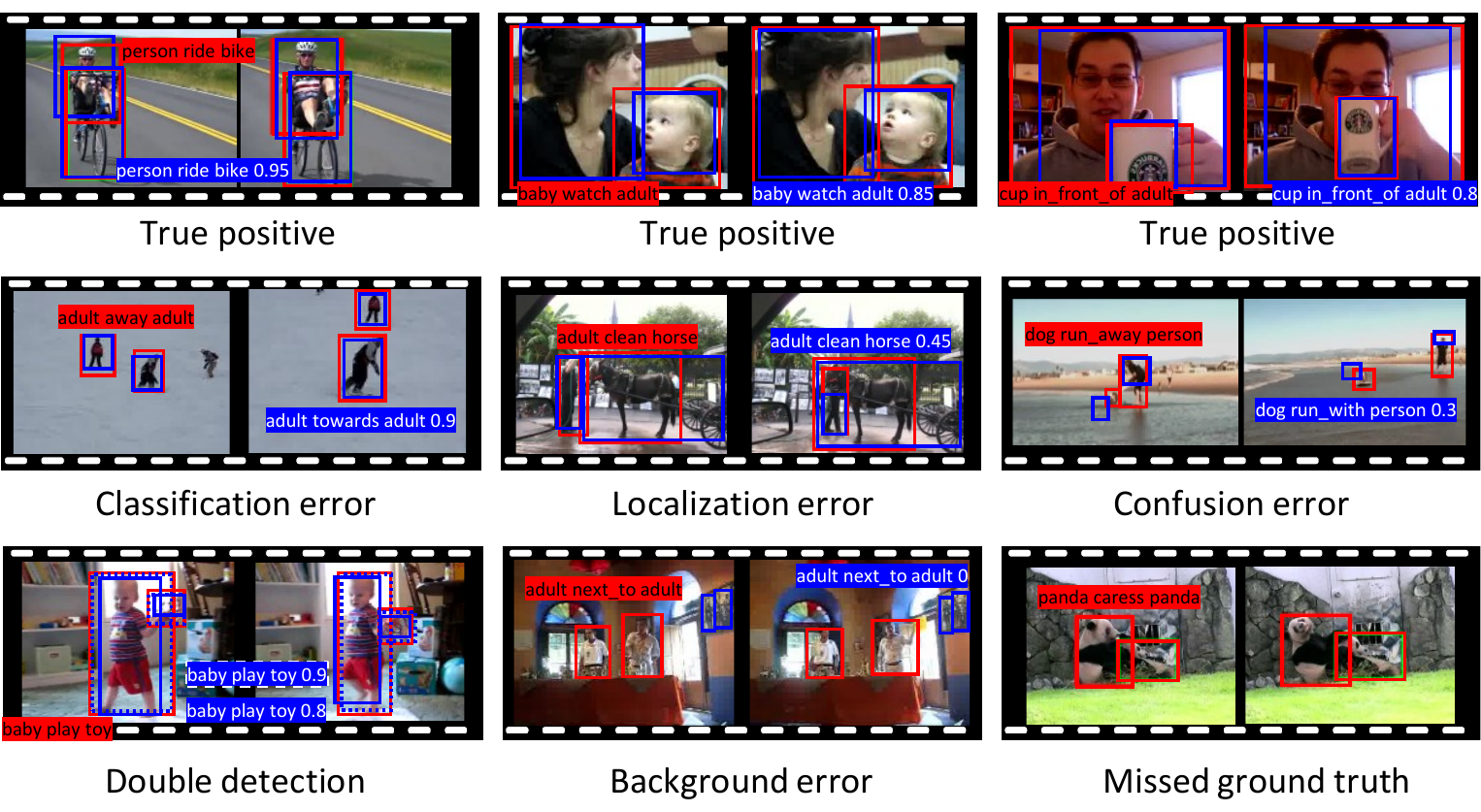}
    \caption{Video relation detection examples of true positives and the six error types from Table \ref{tab:errortypes}. Red boxes indicate ground truth and blue boxes specify predictions. The number in the blue box is the vIoU between the detection and the ground truth. The dashed boxes in double detection represent the best mapped prediction to this ground truth.
    To match a prediction to a ground truth is difficult and many factors could influence the final performance.
    }
    \label{fig:example}
\end{figure}

While calculating the score, we only consider the top-200 predictions for each video. After we get AP for each video, we finally calculate the mean AP (mAP) over all testing/validation videos. The above criteria make it hard for the ground truth to match the prediction.
In this work, we are not only interested in the matches, but also in analyzing the mismatches. In Table~\ref{tab:errortypes}, we have outlined six possible error types, five False Positives, and one False Negative.  We visualize and show qualitative examples of true positives as well as different error types in Figure~\ref{fig:example}. We will use these error types to investigate common pitfalls in current video relation detection approaches.

\subsection{Algorithms under investigation}
We exemplify the use of our diagnostic tool by studying  two state-of-the-art approaches which have conducted experiments on ImageNet-VidVRD and VidOR. Both methods tackled the problem in a three-stage manner, similar to \cite{shang2017video}. However, there are design diﬀerences in each stage which are relevant to highlight.

Liu~\etal~\cite{liu2020beyond} avoid the need to split videos into snippets. In a first stage they generate object tubelets for the whole videos. The second stage refines the tubelet-features and finds relevant object pairs using a graph convolutional network. The third stage focuses on predicting the predicates between related pairs. In this manner, interactions can be detected without a need for snippet splitting.

Su~\etal~\cite{su2020video} is based on the three-stage architecture proposed in Shang~\etal~\cite{shang2017video}. A video is first split into short snippets and subject/object tubelets are generated per snippet. Then, short-term relations are predicted for each tubelet. In the second stage, spatio-temporal features of each pair of object tubelets are extracted and used to predict short-term relation candidates. In the third stage, they maintain multiple relation hypotheses during the association process to accommodate for inaccurate or missing proposals in the earlier steps.

\section{Findings}
In this section, we demonstrate the generality and usefulness of our analysis toolbox by exploring what restricts the performance of video relation detection approaches.
We first conduct a false positive analysis, composed of the first five error types defined in Table~\ref{tab:errortypes} (classification, localization, confusion, double detection, background). Then, we analyze the false negatives, \ie missed ground truth (Miss), along with different relation characteristics that correlate with the false negatives. Finally, we contribute the mAP gain of each error type.

\begin{figure}[t!]
\captionsetup{aboveskip=5pt, belowskip=-5pt}
    \centering
    \includegraphics[width=\textwidth]{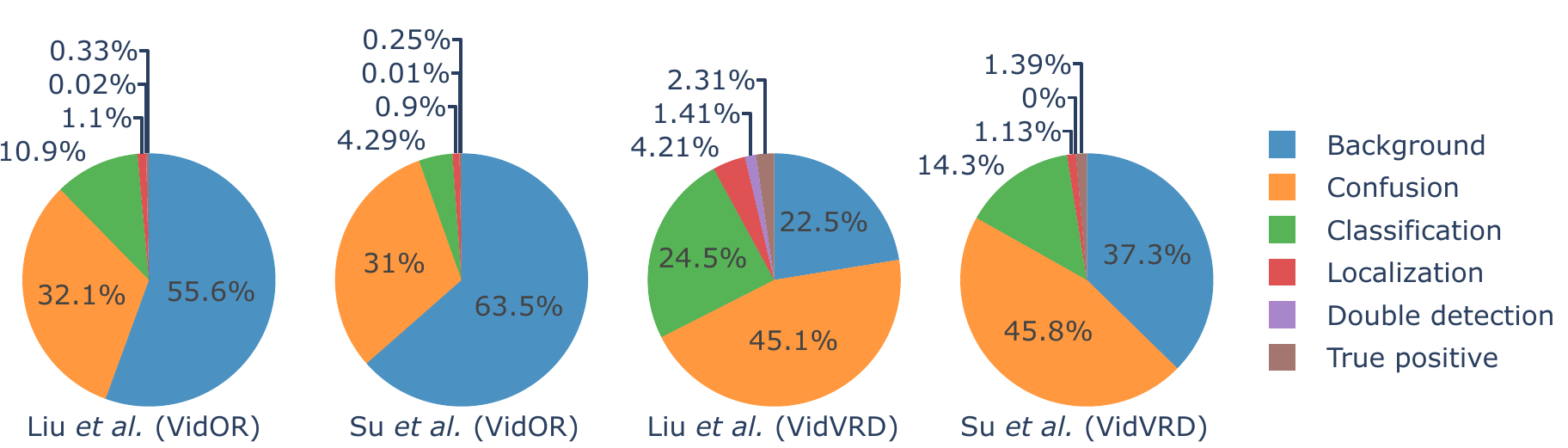}
        \caption{The false positive error breakdown in Liu~\etal~\cite{liu2020beyond} and Su~\etal~\cite{su2020video} on the VidOR and ImageNet-VidVRD datasets. 
        The classification error, which is also one cause of confusion error, as well as background error, should be solved first in future research.
        }
        \label{fig:fp_all}
\end{figure}

\subsection{False positive analysis}
The first experiment investigates which error types are prevalent in current approaches. To answer this question, we break down the false positives and present the distribution of errors for Liu~\etal~\cite{liu2020beyond} and Su~\etal~\cite{su2020video} on ImageNet-VidVRD and VidOR in Figure~\ref{fig:fp_all}. To our surprise we find that in all four cases, the localization error takes only a small part of all false positive in the spatio-temporal detection task. Since in diagnostic papers on well-established detection tasks such as object detection \cite{hoiem2012diagnosing, bolya_2020eccvtide} and temporal action detection \cite{alwassel2018diagnosing}, localization error is important and takes a much larger ratio. 
Due to the large amount of possible triplet combinations, it is more common to have both low overlapping volumes as well as wrong triplet labels, categorized as confusion error. Next, we see that there is almost no double detection error. When predicting predicates, Liu~\etal and Su~\etal keep the top 20 prediction results for each subject-object pair. Thus, the diversity in the predicted detection results make it difficult to map the multiple detections to the same ground truth.

\textbf{Comparison across methods.}
From Figure~\ref{fig:fp_all} we can observe that the background error ratio is much lower in Liu~\etal compared to Su~\etal. Liu~\etal generate less detections where no interesting relations are involved. We attribute this to their proposal generation and filtering stages. Su~\etal's split and merge pipeline might be unable to remove bad proposals efficiently. Another observation is that Liu~\etal's classification error is much higher than the one of Su~\etal on ImageNet-VidVRD. Su~\etal's multiple hypothesis association enables to connect neighbour segments with low predicate prediction scores. When ranking detection results, the scoring reflects the reliability of forming the corresponding hypothesis video relation, enabling a more robust ranking for those with a lower predicate prediction score.  This is beneficial especially for ImageNet-VidVRD with more predicate categories but less training data, resulting in undistinguished classification scores for predicates. Su~\etal have fewer true positives than Liu~\etal, but higher mAP. This also shows that Su~\etal's scoring algorithm outperforms Liu~\etal. In VidOR, with more training data and fewer predicate categories, Su~\etal have a lower classification error ratio than Liu~\etal, but the gap is not as large as on ImageNet-VidVRD. We conclude that Liu~\etal and Su~\etal have their own advantages for dealing with different error types. Both have in common that the background error and classification error should have higher priority than the other error types to gain the most in performance.

\subsection{False negative analysis}
So far, we have only considered the types of false positive errors introduced by the detection algorithms. However, false negative errors (missed ground truth) also influence the mAP. 

\begin{figure}[t!]
\captionsetup{aboveskip=5pt}
\captionsetup[subfigure]{aboveskip=0pt}
     \centering
     \begin{subfigure}[b]{0.24\textwidth}
         \centering
         \includegraphics[width=\textwidth]{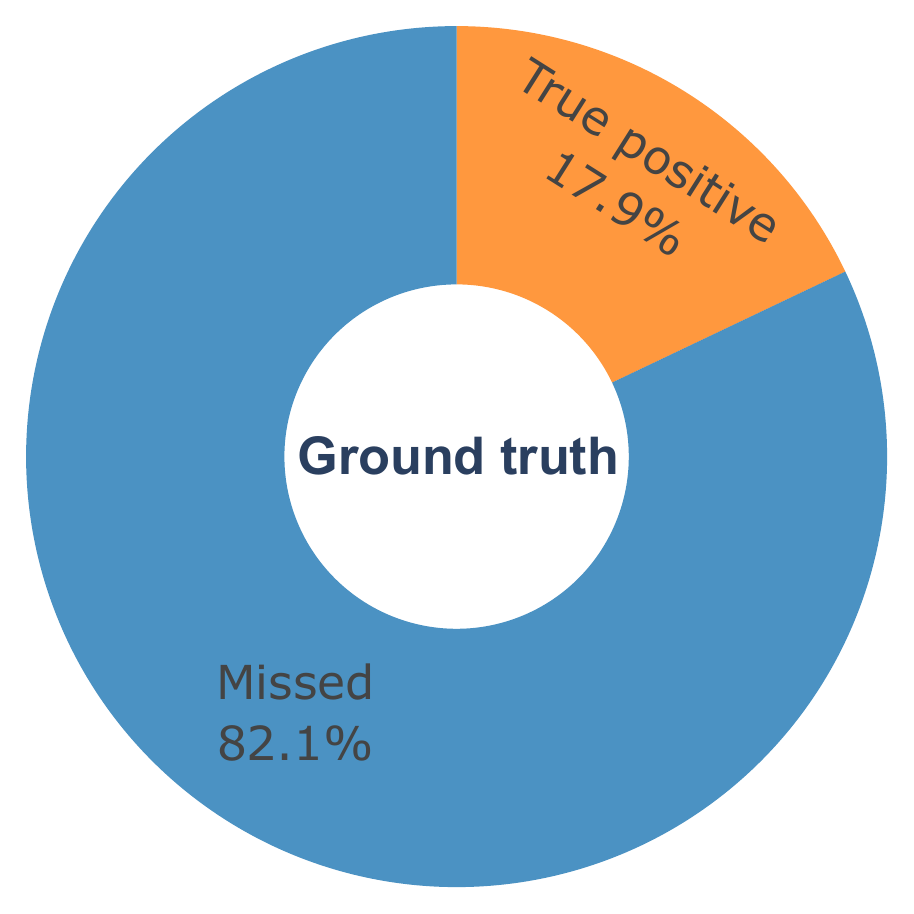}
         \caption{Liu~\etal (VidVRD)}
         \label{fig:liu_ImageNet-VidVRD_fnr}
     \end{subfigure}
     \begin{subfigure}[b]{0.24\textwidth}
         \centering
         \includegraphics[width=\textwidth]{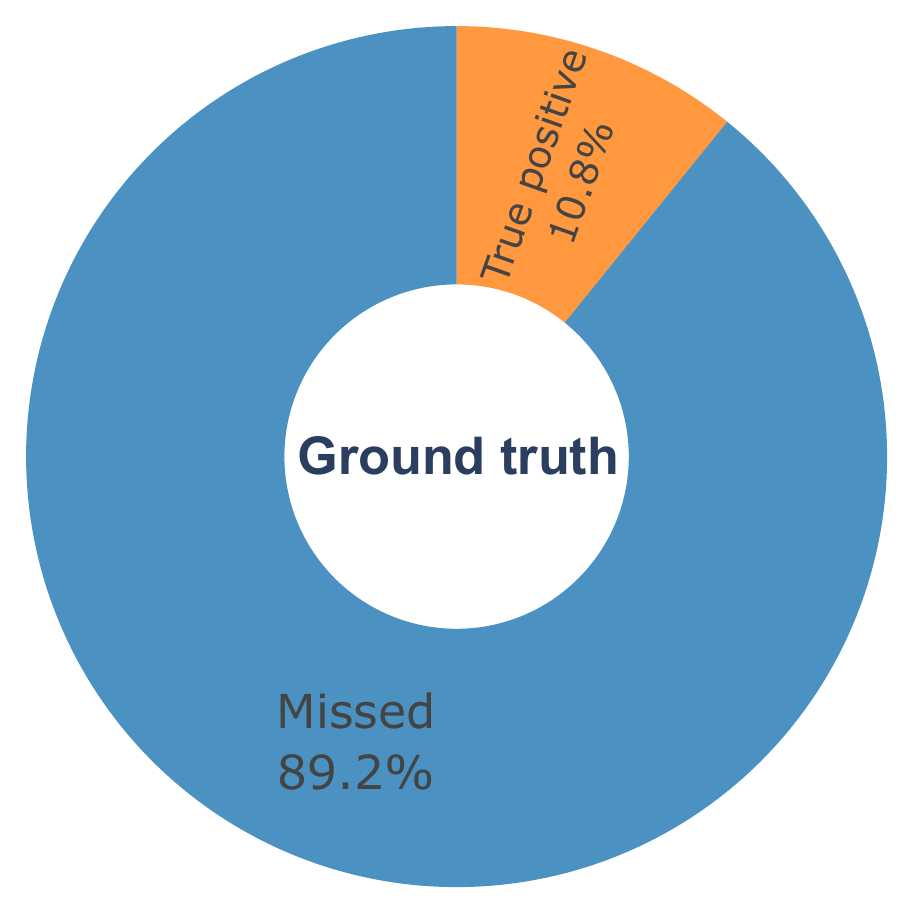}
         \caption{Su~\etal (VidVRD)}
         \label{fig:su_ImageNet-VidVRD_fn}
     \end{subfigure}
     \begin{subfigure}[b]{0.24\textwidth}
         \centering
         \includegraphics[width=\textwidth]{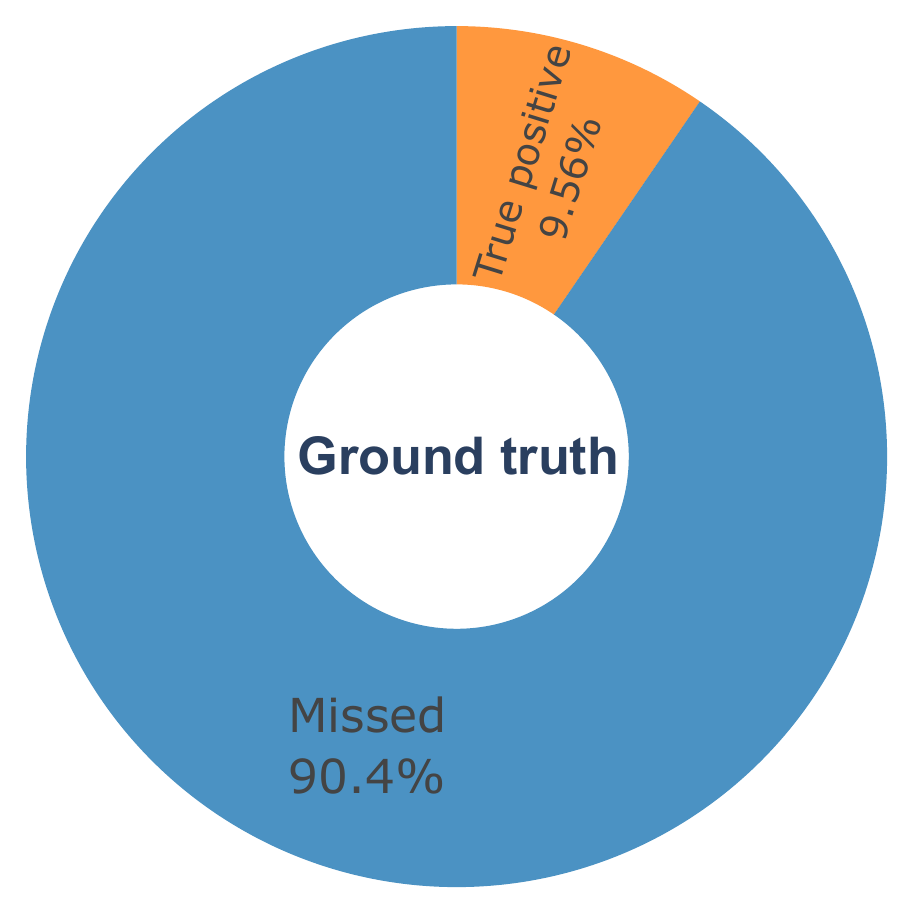}
         \caption{Liu~\etal (VidOR)}
         \label{fig:liu_vidor_fn}
     \end{subfigure}
    \begin{subfigure}[b]{0.24\textwidth}
         \centering
         \includegraphics[width=\textwidth]{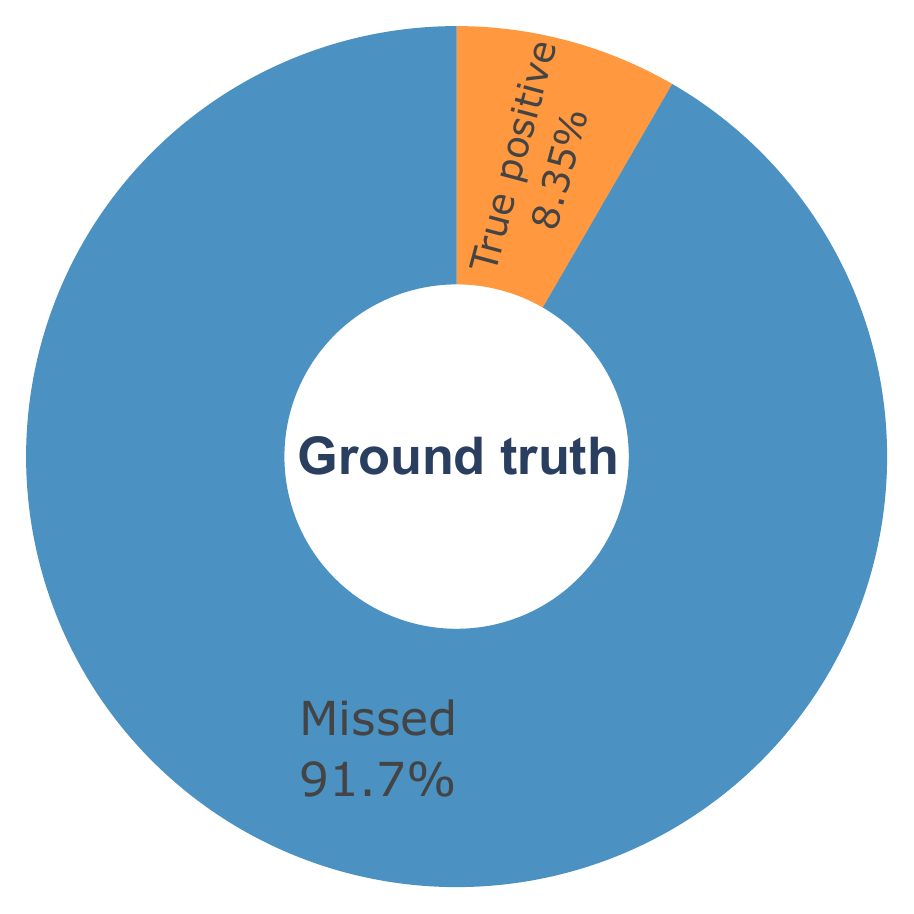}
         \caption{Su~\etal (VidOR)}
         \label{fig:su_vidor_fn}
     \end{subfigure}
    \caption{The missed ground truth error (false positive) ratio on ground truth in Liu~\etal~\cite{liu2020beyond} and Su~\etal~\cite{su2020video} on ImageNet-VidVRD and VidOR datasets. Both have many ground truths undetected.}
    \label{fig:fn_all}
\end{figure}

In Figure~\ref{fig:fn_all} we present the missed ground truth ratios for Liu~\etal and Su~\etal on ImageNet-VidVRD and VidOR. 
For both ImageNet-VidVRD and VidOR, roughly 90\% of the ground truth relation instances remain undetected. VidOR has a higher missed ground truth ratio, highlighting the more complex nature of the dataset. On ImageNet-VidVRD, Liu~\etal detect more instances than Su~\etal but attribute them with lower scores, leading to a lower mAP value. This tells us that proposal-based methods can cover more relations, while Su~\etal's scoring method helps to better rank detected predictions.
It is insightful to study what makes these missed ground truth instances difficult to detect. Towards this end, we group the instances according to six relation characteristics defined below:

\begin{itemize}[noitemsep]
\item \textbf{Length:} we measure relation length by the duration in seconds and create three different length groups: Short (S: (0, 10]), Medium (M: (10, 20]), and Long (L: > 20). Overall, most of the instances are short, both in ImageNet-VidVRD (94.11\%) and VidOR (80.06\%). The number of medium and long relations are roughly similar.

\item \textbf{Number of predicate instances}: we count the total number of predicate instances over all videos and create four categories: XS: (0, 10]; S: (10, 100]; M: (100, 1000]; L: (1000, 10000]; XL: (10000, 100000]; XXL: >100000.
\item \textbf{Number of subject instances}: idem but for subjects.
\item \textbf{Number of object instances}:
idem but for objects.
\item \textbf{Subject pixel scale}: we take the average of the bounding boxes for the subject trajectories and group the mean bounding box. We define subjects with pixel areas between 0 and 162 as extra small (XS), 162 to 322 as small (S), 322 to 962 as medium (M), 962 to 2882 as (L), and 2882 and above as extra large (XL).
\item \textbf{Object pixel scale}:
idem but for objects.
\end{itemize}

\begin{figure}[t!]
\captionsetup{aboveskip=5pt}
\captionsetup[subfigure]{aboveskip=0pt}
\centering
\begin{subfigure}[b]{\linewidth}
     \includegraphics[width=\linewidth]{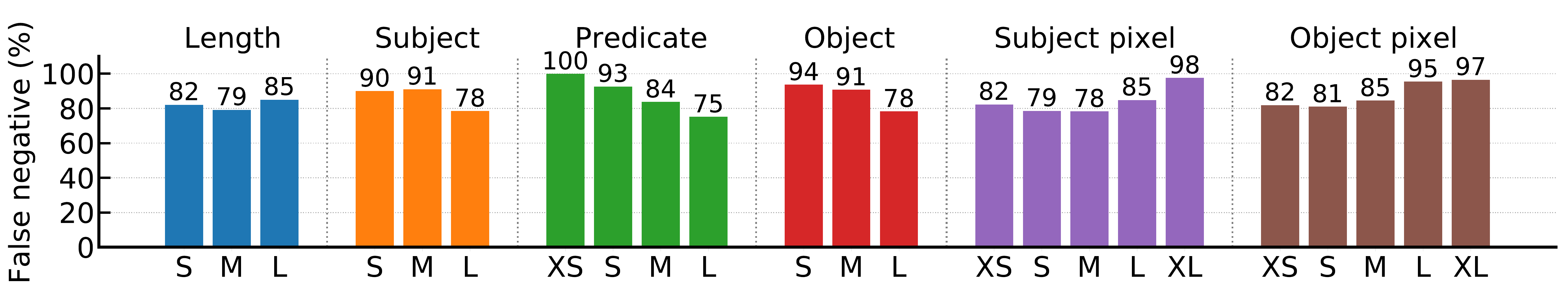}
  \caption{Liu~\etal (ImageNet-VidVRD).}
\label{fig:char_liu_ImageNet-VidVRD}
\end{subfigure}
\begin{subfigure}[b]{\linewidth}
\centering
    \includegraphics[width=\linewidth]{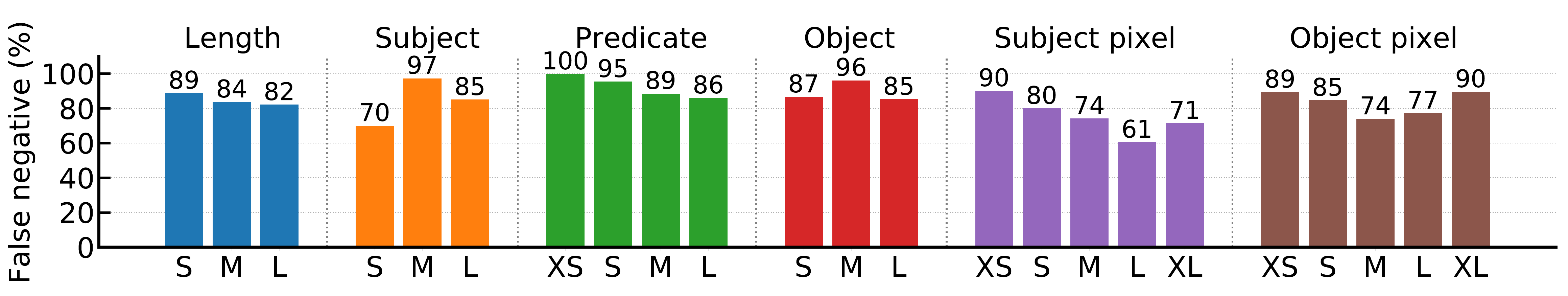}
  \caption{Su~\etal (ImageNet-VidVRD).}
\label{fig:char_su_ImageNet-VidVRD}
\end{subfigure}
\begin{subfigure}[b]{\linewidth}
\centering
    \includegraphics[width=\linewidth]{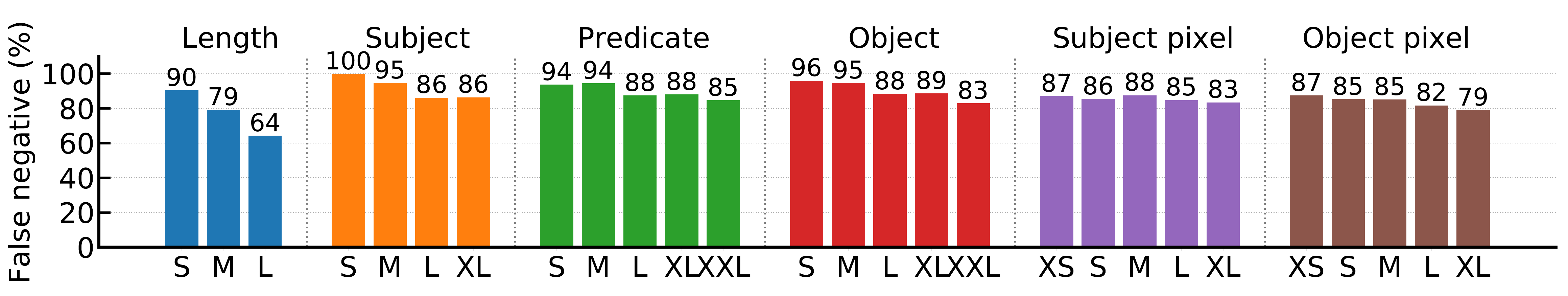}
  \caption{Liu~\etal (VidOR).}
\label{fig:char_liu_vidor}
\end{subfigure}
\begin{subfigure}[b]{\linewidth}
\centering
    \includegraphics[width=\linewidth]{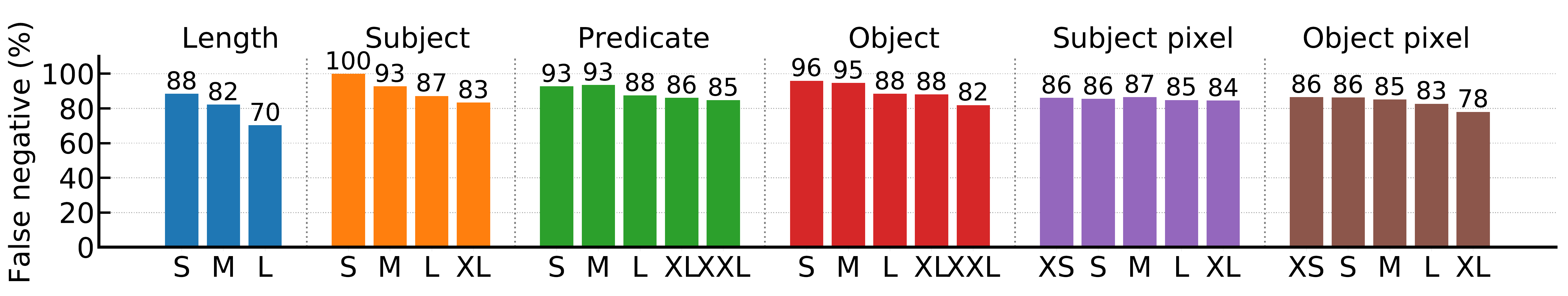}
  \caption{Su~\etal (VidOR).}
\label{fig:char_su_vidor}
\end{subfigure}
\caption{Relation characteristics of Liu~\etal and Su~\etal on ImageNet-VidVRD and VidOR. Relations with fewer subject/predicate/object instances and smaller subject/object pixel areas are more difficult to detect.
}
\label{fig:char_full}
\end{figure}

Figure ~\ref{fig:char_full} shows the overview of the effect for all relation characteristics for both Liu~\etal and Su~\etal on ImageNet-VidVRD and VidOR. We first observe a long-tail issue for the predicates. On ImageNet-VidVRD, both methods completely fail on relation instances for which the predicate category has fewer than 10 samples. This means that datasets with more training samples are essential to this task, or methods should better exploit the few available samples. 
Another observation is that Su~\etal have fewer false negatives on long-range relations on ImageNet-VidVRD, even though Liu~\etal focus on long-range representations in their approach.
This may be due to the construction of the ImageNet-VidVRD dataset,
which was built through asking annotators to label segment-level visual relation instances in decomposed videos. This annotation procedure results in an abundance of relations that can be recognized without the need for long-range information. VidOR is annotated differently. Given a pair of object tubelets, the annotators are asked to find and temporally localize relations, resulting in more long-lasting relations.
The patterns regarding the number of subject and object instances are intuitive in VidOR; the more instances to train on the better. Moreover, subjects and objects with larger size are easier to detect than smaller size. This pattern does, however, not hold for ImageNet-VidVRD, which could be due to the overall dataset size. Since the numbers of ‘XL’ subpxl and ‘XL’ objpxl in ImageNet-VidVRD are much lower than in VidOR.

\begin{figure}[t!]
\captionsetup{aboveskip=5pt}
\captionsetup[subfigure]{aboveskip=0pt}
\centering
\begin{subfigure}[b]{0.95\linewidth}
\centering
     \includegraphics[width=\linewidth]{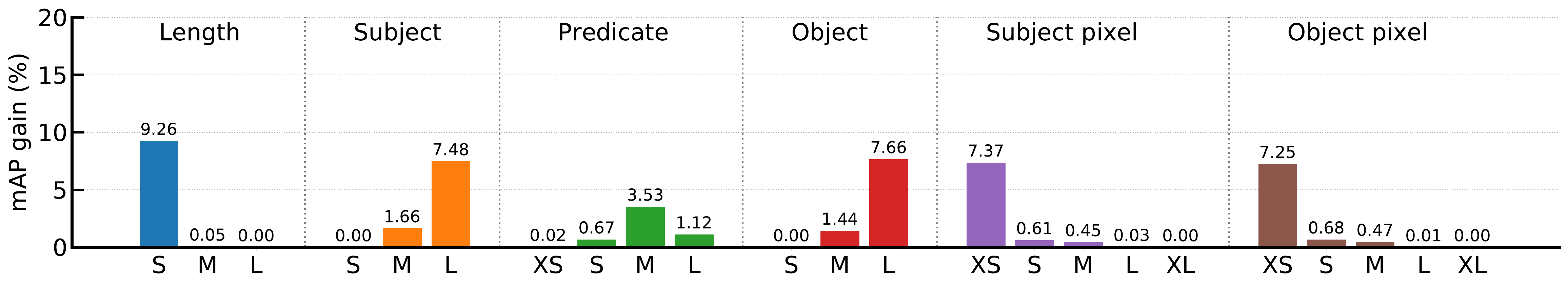}
  \caption{Liu~\etal (ImageNet-VidVRD).}
\label{fig:char_map_liu_ImageNet-VidVRD}
\end{subfigure}
\begin{subfigure}[b]{0.95\linewidth}
\centering
    \includegraphics[width=\linewidth]{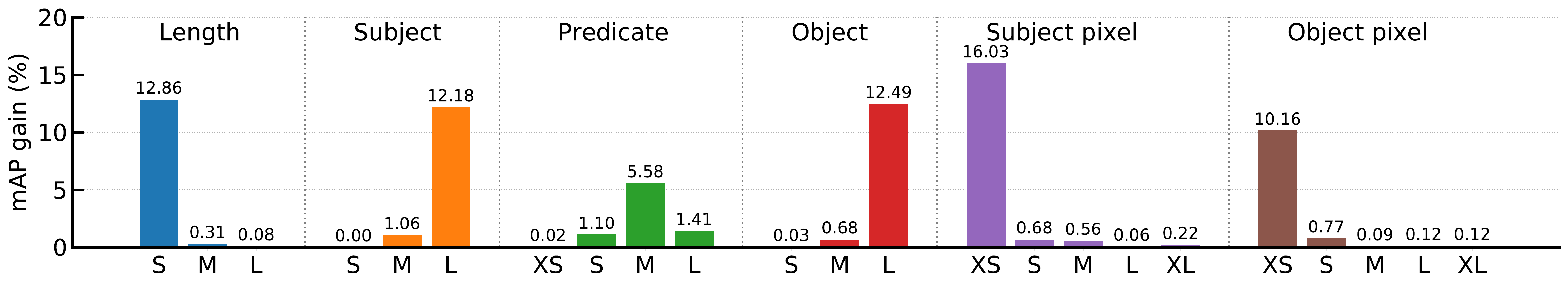}
  \caption{Su~\etal (ImageNet-VidVRD).}
\label{fig:char_map_su_ImageNet-VidVRD}
\end{subfigure}
\begin{subfigure}[b]{0.95\linewidth}
\centering
    \includegraphics[width=\linewidth]{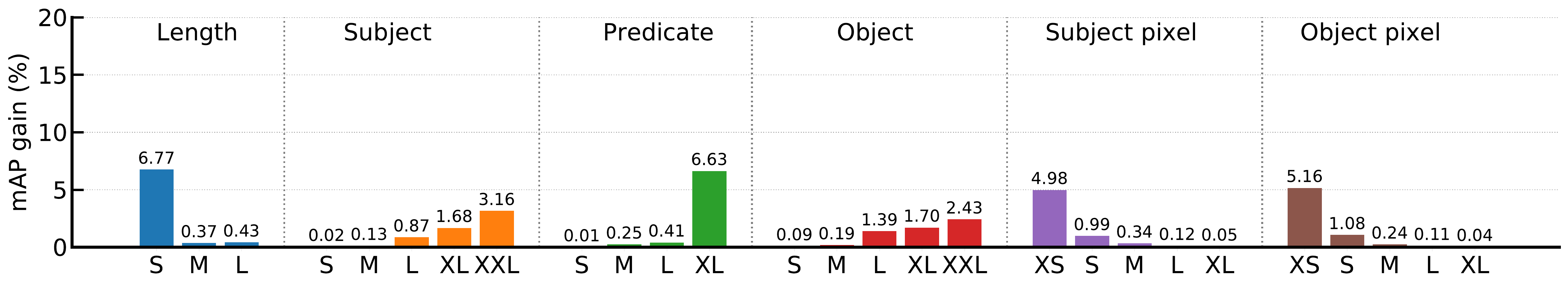}
  \caption{Liu~\etal (VidOR).}
\label{fig:char_map_liu_vidor}
\end{subfigure}
\begin{subfigure}[b]{0.95\linewidth}
\centering
    \includegraphics[width=\linewidth]{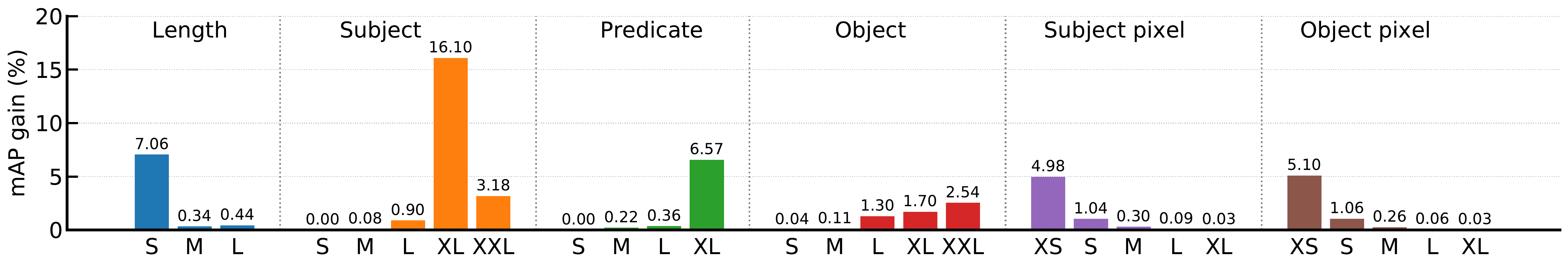}
  \caption{Su~\etal (VidOR).}
\label{fig:char_map_su_vidor}
\end{subfigure}
\caption{The mAP gain on relation characteristics of Liu~\etal and Su~\etal on ImageNet-VidVRD and VidOR. Focusing on detecting relation instances with a short temporal timespan, a large number of instances, and small pixel areas for the subject and object will improve the mAP by the largest margin. 
}
\label{fig:char_map}
\end{figure}

To deepen the analysis of each characteristic's effect, we calculate the mAP gain after dropping the missed ground truths under this characteristic. From Figure~\ref{fig:char_map}, we observe that not all characteristics contribute equally to gains in mAP. It reveals that to improve the final metric the most, methods should focus on detecting relation instances with a short temporal timespan, a large number of instances, and small pixel areas for the subject and object.

\subsection{mAP sensitivity}
Where we have so far looked into which errors are most prevalent, we also want to examine to what extent each error type in Table~\ref{tab:errortypes} is holding back progress. We do so by quantifying the impact on the mAP for each error type by means of an oracle fix.
We show how the mAP changes when each error type would be fixed. Rather than only removing the predictions causing this error~\cite{alwassel2018diagnosing}, we define the following cures for each of the main error types:
\begin{itemize}[noitemsep]
\itemsep0em
    \item \textbf{Classification cure}: Correct the class of the detection (thereby making it a true positive). If this results in a duplicate detection, remove the lower scoring detection.
    \item \textbf{Localization cure}: Set the localization of the detection equal to the ground truth localization (thereby making it a true positive). If this results in a duplicate detection, remove the lower scoring detection.
    \item \textbf{Confusion cure}: Since we cannot be sure of which ground truth the detector was attempting to match to, we remove the false positive detection. 
    \item \textbf{Double detection cure}: Remove the duplicate detection with lower score.
    \item \textbf{Background cure}: Remove the background detection. 
    \item \textbf{Missed ground truth cure}: Reduce the number of ground truth instances in the mAP calculation by the number of missed ground truth.
\end{itemize}

\begin{figure}[t]
\captionsetup{aboveskip=5pt,belowskip=-10pt}
     \centering
    \includegraphics[width=\textwidth]{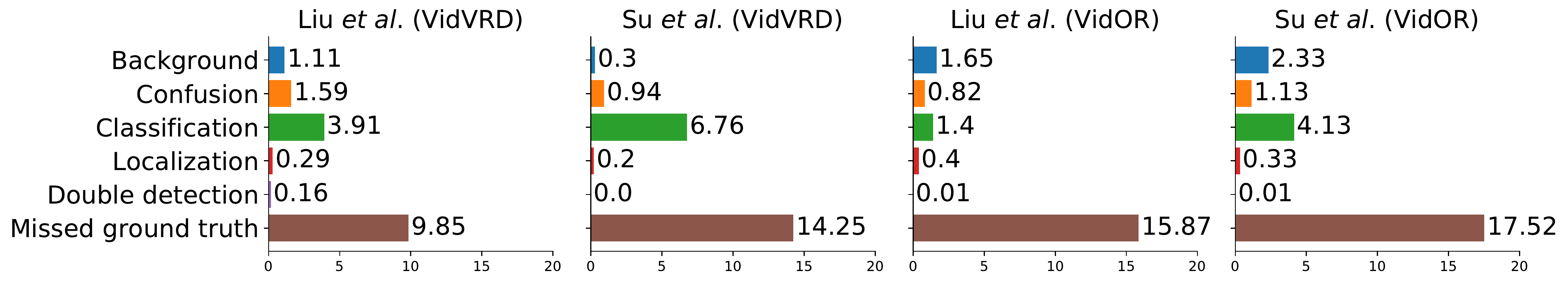}
    \caption{The mAP gain in Liu~\etal~\cite{liu2020beyond} and Su~\etal~\cite{su2020video} on ImageNet-VidVRD and VidOR. Fixing missed ground truth error will maximize the performance improvement.
    }
    \label{fig:map_gain}
\end{figure}

Figure~\ref{fig:map_gain} shows the error types impact on the mAP. Note that the sum of each error type's mAP gain is not 100\%. The reason is due to the property of mAP. If we fix the error types progressively, the final mAP will be 100\%. But the later fixed error types will gain more weights than earlier fixed error types. For a meaningful comparison, we fix them separately. In Figure~\ref{fig:map_gain}, fixing missed ground truth errors will improve the mAP by a large margin, Su \etal with 14.25\% on ImageNet-VidVRD and 17.52\% on VidOR. However, in practice, we cannot simply drop these missed ground truths. The solution is to include more ground truths in the selected top 200 detections of a video. And many detections that could be matched to missed ground truths are not selected due to their low scores. We believe one direction is improving the performance of the predicate prediction module, to give the background proposals low scores and proposals of correct predicate categories high scores. 
This will also fix the classification errors and background errors to boost the final mAP further.

\section{Conclusion}
This work performs a series of analyses to understand the challenging problem of video relation detection better. Using two canonical approaches, we first perform false positive analyses and define the different types of errors. Two error types are prevalent across approaches and datasets: confusion with non-matching ground truth relations and detecting relations that are part of the background. We then perform false negative analyses, which show that most ground truth instances are missed entirely. Focusing on detecting relation instances with a short temporal length, a large number of instances, and small pixel areas for the subject and object will improve the mAP the most. Lastly, to create a future outlook, we investigate several cures for common errors and find that the ability to discard background relations provides the shortest path to improve video relation detection performance. Our toolbox is generic and can be employed on top of any video relation detection approach. We make the toolbox and evaluation scripts publicly available to help researchers dissect their video relation detection approaches. Currently our tool only consider the single variant's effect to the final metric, we will investigate a multivariate statistical analysis in the future.

\bibliography{egbib}
\end{document}


\maketitle

{
\small  
\begin{longtable}{clcr}
\toprule
Super Category & \multicolumn{1}{c}{Category} & Classes & \multicolumn{1}{c}{Instances} \\
\toprule
\multicolumn{4}{c}{Subjects \& Objects} \\
\midrule
Animal &  \multicolumn{1}{p{5cm}}{turtle, antelope, lion, cattle, red\_panda, horse,  monkey, fox, elephant, bird, sheep, giant\_panda, squirrel, bear, tiger, snake, rabbit, whale, dog, domestic\_cat, lizard, hamster, zebraf} & 23 & 39097  (63.57\%) \\
Furniture & sofa & 1 & 356 \,  (0.58\%) \\
Person & person & 1 & 8536 (13.88\%) \\
Sports & ball, frisbee, skateboard & 3 & 519 \, (0.84\%) \\
Vehicle &\multicolumn{1}{p{5cm}}{bicycle, motorcycle, airplane, watercraft, bus, train, car} & 7 & 12996 (21.13\%) \\
\midrule
\multicolumn{4}{c}{Predicates} \\
\midrule
Action & bite, chase, drive  & 14 & 2956 \, (9.61\%) \\
Spatial & \multicolumn{1}{p{5cm}}{walk\_above, stand\_behind, next\_to, fly\_toward, sit\_right, jump\_with, walk\_behind, creep\_above, stand\_front, run\_front, run\_left, jump\_next\_to, right, creep\_right, walk\_left, fly\_left, swim\_beneath, swim\_behind, creep\_left, creep\_away, creep\_next\_to, lie\_left, creep\_behind, walk\_right, stand\_inside, stand\_left, jump\_above, move\_past, run\_past, walk\_toward, left, creep\_toward, jump\_toward, walk\_next\_to, sit\_inside, stand\_right, run\_next\_to, lie\_behind, fly\_right, lie\_beneath, sit\_left, past, run\_away, stop\_above, move\_with, move\_right, lie\_above, stop\_with, jump\_left, stop\_right, front, jump\_beneath, walk\_past, sit\_behind, move\_above, lie\_next\_to, walk\_beneath, walk\_with, move\_beneath, run\_above, run\_with, toward, run\_beneath, stop\_behind, jump\_behind, move\_left, walk\_front, move\_toward, move\_behind, above, move\_away, swim\_left, stand\_with, stop\_left, stand\_beneath, beneath, stand\_next\_to, swim\_front, creep\_beneath, lie\_front, move\_front, fly\_above, sit\_beneath, sit\_front, jump\_away, stop\_beneath, sit\_above, run\_behind, fly\_front, creep\_front, faster, stop\_next\_to, away, lie\_with, run\_toward, lie\_right, lie\_inside, stop\_front, run\_right, taller, stand\_above, swim\_with, jump\_past, fly\_away, creep\_past, walk\_away, behind, move\_next\_to, jump\_front, swim\_next\_to, jump\_right, swim\_right, sit\_next\_to, fly\_with, larger, fly\_behind, fly\_next\_to, fly\_past}
 & 118 & 27796 (90.39\%) \\
\bottomrule
\caption{Subject/object and relation categories in Imagenet-VidVRD dataset, organized by super categories. Note the bias towards animals and spatial relations. }
\label{tab:vidvrd_type}
\end{longtable}

\pagebreak

\begin{longtable}{clcr}
\toprule
 Super Category & \multicolumn{1}{c}{Category} & Classes & \multicolumn{1}{c}{Instances}\\
\toprule
\multicolumn{4}{c}{Subjects \& Objects} \\
\midrule
Accessory & handbag, backpack, suitcase & 3 & 5948 \, (1.00\%) \\
Animal & \multicolumn{1}{p{5cm}}{leopard, snake, chicken, hamster/rat, stingray, antelope, turtle, panda, tiger, sheep/goat, crocodile, pig, fish, cat, dog, lion, bird, elephant, duck, camel, kangaroo, crab, cattle/cow, penguin, horse, squirrel, bear, rabbit} & 28 & 51009 \, (8.58\%) \\
Appliance & \multicolumn{1}{p{5cm}}{refrigerator, sink, oven, microwave, electric\_fan} & 5 & 2034  \, (0.34\%) \\
Electronic & \multicolumn{1}{p{5cm}}{camera, cellphone, screen/monitor, laptop} & 4 & 16005  \, (2.69\%) \\
Food & \multicolumn{1}{p{5cm}}{fruits, bread, cake, vegetables, dish} & 5 & 8094  \, (1.36\%) \\
Furniture & \multicolumn{1}{p{5cm}}{table, toilet, stool, chair, sofa} & 5 & 41089 \,  (6.91\%) \\
Indoor & toy & 1 & 30034  \, (5.05\%) \\
Kitchen & bottle, cup & 2 & 21330 \, (3.59\%) \\
Other & \multicolumn{1}{p{5cm}}{piano, baby\_walker, baby\_seat, faucet, guitar} & 5 & 15789 \, (2.65\%) \\
Outdoor & bench, stop\_sign, traffic\_light & 3 & 1744 \, (0.29\%) \\ [1mm]
Person & adult, baby, child & 3 & 368549 (61.97\%) \\
Sports & \multicolumn{1}{p{5cm}}{bat, snowboard, surfboard, skateboard, racket, ball/sports\_ball, ski, frisbee} & 8 & 16697 \, (2.81\%) \\ [1mm]
Vehicle & \multicolumn{1}{p{5cm}}{scooter, watercraft, bus/truck, bicycle, aircraft, car, motorcycle, train} & 8 & 16382  \, (2.75\%) \\
\midrule
\multicolumn{4}{c}{Predicates} \\
\midrule
Action & \multicolumn{1}{p{5cm}}{kiss, bite, push, point\_to, wave\_hand\_to, drive, carry, open, watch, throw, clean, feed, wave, shake\_hand\_with, play(instrument), get\_off, hug, touch, hold, pat, press, chase, close, release, grab, lift, smell, hold\_hand\_of, knock, lick, cut, kick, pull, get\_on, lean\_on, hit, speak\_to, ride, shout\_at, squeeze, caress, use} & 42 & 69066 (23.23\%) \\
Spatial & \multicolumn{1}{p{5cm}}{above, next\_to, beneath, in\_front\_of, away, behind, inside, towards} & 8 & 228286 (76.77\%) \\
\bottomrule
\caption{Subject/object and relation categories in VidOR, organized by their super categories. Note the bias towards persons and spatial relations.}
\label{tab:vidor_type}
\end{longtable}
}